\documentclass[11pt]{article}

\usepackage[final]{acl}

\usepackage{times}
\usepackage{latexsym}
\usepackage{amsmath}
\usepackage{makecell}
\usepackage{booktabs}

\usepackage[T1]{fontenc}

\usepackage[utf8]{inputenc}

\usepackage{microtype}

\usepackage{inconsolata}

\usepackage{graphicx}

\title{SampoNLP: A Self-Referential Toolkit for Morphological Analysis of Subword Tokenizers}

\author{
\textbf{Iaroslav Chelombitko} \\
DataSpike, aglabx \\
Neapolis University Pafos \\
Paphos, Cyprus \\
\texttt{i.chelombitko@nup.ac.cy}
\And
\textbf{Ekaterina Chelombitko} \\
DataSpike \\
Dubai, UAE \\
\texttt{ekaterina@dataspike.io}
\And
\textbf{Aleksey Komissarov} \\
aglabx \\
Paphos, Cyprus \\
\texttt{ad3002@gmail.com}
}

\begin{document}
\maketitle
\begin{abstract}
The quality of subword tokenization is critical for Large Language Models, yet evaluating tokenizers for morphologically rich Uralic languages is hampered by the lack of clean morpheme lexicons. 

We introduce SampoNLP, a corpus-free toolkit for morphological lexicon creation using MDL-inspired Self-Referential Atomicity Scoring, which filters composite forms through internal structural cues - suited for low-resource settings.

Using the high-purity lexicons generated by SampoNLP for Finnish, Hungarian, and Estonian, we conduct a systematic evaluation of BPE tokenizers across a range of vocabulary sizes (8k–256k). We propose a unified metric, the Integrated Performance Score (IPS), to navigate the trade-off between morpheme coverage and over-splitting. By analyzing the IPS curves, we identify the "elbow points" of diminishing returns and provide the first empirically grounded recommendations for optimal vocabulary sizes (k) in these languages. Our study not only offers practical guidance but also quantitatively demonstrates the limitations of standard BPE for highly agglutinative languages. The SampoNLP library and all generated resources are made publicly available\footnote{\url{https://github.com/AragonerUA/SampoNLP}}.
\end{abstract}

\section{Introduction}

The performance of subword tokenization algorithms like Byte-Pair Encoding (BPE) \cite{sennrich-etal-2016-neural} is a cornerstone of modern Natural Language Processing (NLP). While highly effective for many languages, their purely statistical nature poses a significant challenge for morphologically rich, agglutinative languages \cite{bostrom-durrett-2020-byte, rust-etal-2021-good}. In the Uralic family, a group of languages known for its complex morphology and diverse linguistic phenomena \cite{riessler-2022-uralic}, words are often long concatenations of morphemes (e.g., Finnish talo-i-ssa-ni-ko-kaan - "not in my houses either?"). For such languages, the quality of tokenization is not just an engineering detail but a critical factor that determines a model's ability to grasp grammatical structure and generalize effectively \cite{hamalainen-2021-neural, gerz-etal-2018-language}. This raises a pressing, yet under-explored, practical question, known to be a challenge in Uralic NLP: What is the optimal tokenizer vocabulary size (k) to achieve robust morphological representation? The importance of this question was highlighted by recent work demonstrating the benefits of specialized tokenizers for these languages \cite{chelombitko-komissarov-2024-specialized}.

Addressing this question reveals a more fundamental problem: the scarcity of high-purity morphological resources for evaluation. While lexical data is available in spell-checking dictionaries, their raw combination of stems and affixes results in a noisy candidate list. Manual curation is not scalable, and established corpus-based methods like Morfessor \cite{creutz-lagus-2007-unsupervised} are ill-suited for the many low-resource Uralic languages \cite{arkhangelskiy-2020-corpus}.

To address this challenge, we present SampoNLP, a toolkit based on a corpus-free and self-referential pipeline for refining morphological lexicons. The proposed method, "MDL-inspired Self-Referential Atomicity Scoring," draws its theoretical motivation from the Minimum Description Length principle \cite{rissanen-1978-modeling}, but adapts it to a type-only setting. The core algorithm iteratively estimates the atomicity of each candidate, distinguishing between simple and composite forms by analyzing internal structural patterns within the dataset itself. This lightweight and reproducible approach offers a practical way to produce cleaner morphological resources, a recognized need for data-scarce environments where traditional corpus-based methods are not viable \cite{riessler-2022-uralic}.

Having established a robust methodology for resource creation, we leverage our generated lexicons to address the core problem of this paper: the vocabulary-morphology trade-off inherent in BPE tokenization \cite{bostrom-durrett-2020-byte}. We conducted a systematic evaluation of BPE tokenizers for Finnish, Hungarian, and Estonian across vocabulary sizes from 8k to 256k. The development of novel evaluation frameworks that go beyond downstream performance is a growing area of research \cite{chelombitko2024qtok}. In line with this, to precisely navigate the aforementioned trade-off, we introduce the Integrated Performance Score (IPS), a single metric that balances Lexical Morpheme Coverage (LMC) against the Over-Split Rate (OSR). This allows us to model the performance curve and identify the optimal vocabulary range, providing a principled answer to our central research question.

Our contributions are thus twofold and equally significant:

\begin{enumerate}
    \item \textbf{A Corpus-Free Morphological Method:} We introduce a fully automatic and reproducible pipeline for refining morphological lexicons without relying on corpus frequencies or external resources, released as an open-source toolkit, \textit{SampoNLP}.
    \item \textbf{A Quantitative Evaluation:} We conduct a systematic analysis of BPE tokenizers for Finnish, Estonian, and Hungarian, examining how vocabulary size affects morphological granularity through newly defined metrics of coverage and over-segmentation.
\end{enumerate}

\section{Related Work}

The evaluation and optimization of subword tokenization for morphologically rich languages intersects several research areas: subword tokenization algorithms, unsupervised morphological analysis, rule-based analyzers, and language-specific NLP for Uralic languages.

\subsection{Subword Tokenization and Morphology}

Byte-Pair Encoding (BPE) \cite{sennrich-etal-2016-neural} has become the de facto standard for subword tokenization in modern NLP. Alongside it, methods like the Unigram Language Model \cite{kudo-2018-subword} have been proposed, but the purely statistical nature of these approaches presents well-documented challenges for morphologically complex languages. The work of \cite{bostrom-durrett-2020-byte} demonstrated that BPE tokenizers often fail to align with linguistic morpheme boundaries. Interestingly, parallel challenges in identifying meaningful subsequence units have been explored in domains beyond NLP, such as the tokenization of biological sequences like primate genomes \cite{popova2025repeatsdrivevocabularybytepair}.

The question of optimal vocabulary size has often been guided by heuristics or evaluated indirectly via downstream task performance \cite{mielke-etal-2021-results}. Our work directly addresses this gap by proposing a methodology for intrinsic, morphologically-grounded evaluation to provide data-driven recommendations for Uralic languages.

\subsection{Unsupervised Morphological Analysis}

The unsupervised discovery of morphological structure has a rich history. One major family of approaches relies on statistical cues from corpora to identify boundaries. Classic methods such as Branching Entropy and Accessor Variety \cite{feng2004accessor} analyze the predictability of subsequent characters to hypothesize morpheme breaks.
Another prominent family of methods is based on the Minimum Description Length (MDL) principle. Morfessor \cite{creutz-lagus-2007-unsupervised} and its variants represent the canonical probabilistic approach, finding a lexicon that best compresses a text corpus. While successful, these methods are fundamentally corpus-based, requiring token frequency information that may not be available in low-resource settings.

Our approach, while MDL-inspired, operates in a corpus-free, type-only regime. It represents a different paradigm: self-referential filtering of a candidate list. By operating purely on the internal structure of a candidate set, we provide a lightweight method suited to resource-scarce scenarios, a persistent challenge in Uralic NLP \cite{arkhangelskiy-2020-corpus}.

\subsection{Rule-Based Analyzers and Tokenization for Uralic Languages}

For Uralic languages, rule-based morphological analyzers built on Finite-State Transducers (FSTs) like Omorfi \cite{pirinen-2015-omorfi} and the GiellaLT\footnote{\url{https://giellalt.github.io/}} infrastructure \cite{trosterud-2012-uralic} are invaluable resources. While their generative outputs are linguistically comprehensive, they are not directly optimized for use as a minimal reference morphemes lexicon. Our IMDP pipeline offers a contrasting approach: a data-driven methodology for distilling such a lexicon from a type-only candidate list, as can be extracted from dictionary-based resources like Hunspell, without requiring token frequencies from a corpus.

The challenge of effective tokenization for this language family has recently gained significant attention. Broader findings have established that language-specific modeling is crucial for morphologically rich languages, with studies on Finnish demonstrating clear benefits of monolingual models like FinBERT over multilingual ones \cite{virtanen-etal-2019-multilingual}. Building on this principle, a recent study by \cite{chelombitko-komissarov-2024-specialized} specifically addressed the severe under-representation of Uralic languages in large multilingual models. They demonstrated that training specialized, large-vocabulary monolingual tokenizers yields substantial improvements in compression efficiency. However, while establishing the need for specialized resources, their work left the question of how to determine an optimal vocabulary size open for future investigation.

Concurrently, the need for better evaluation metrics has become a prominent research topic. The Qtok framework \cite{chelombitko2024qtok}, for instance, proposed a comprehensive approach to evaluating multilingual tokenizer quality, while other studies have also advocated for moving beyond downstream task performance towards more intrinsic, linguistically-informed measures \cite{acs-2019-tokenization}. Our Integrated Performance Score (IPS) directly addresses this call from the community for more morphologically-grounded metrics.

Our current work builds on these foundations. It utilizes similar high-quality data sources as those in \cite{chelombitko-komissarov-2024-specialized} to train the tokenizers being evaluated. Furthermore, by proposing a concrete methodology, it answers the call for better evaluation and finds the optimal vocabulary sizes that the former study alluded to, thus providing a logical next step in this line of research.

\section{Methodology. The IMDP Pipeline}

\begin{figure*}[htbp]
    \centering
    \includegraphics[width=1.0\linewidth]{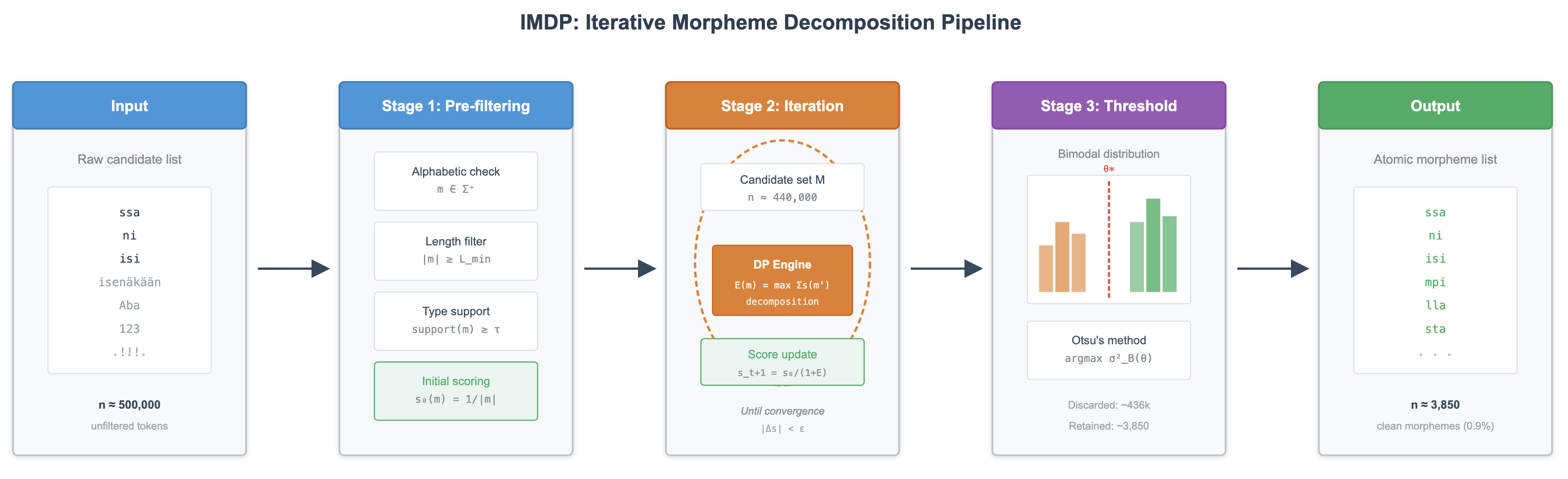}
    \caption{An overview of the Iterative Morphological Decomposition Pipeline (IMDP).}
    \label{fig:imdp_pipeline}
\end{figure*}

To create a high-purity morpheme lexicon from a noisy, raw list of candidate forms, we propose the Iterative Morphological Decomposition Pipeline (IMDP). Our approach is designed to be fully automatic and operates in a corpus-free, type-only regime, requiring only the candidate list as input. The core of the pipeline is a method we term "MDL-inspired Self-Referential Atomicity Scoring," which iteratively evaluates how "fundamental" each candidate is relative to the entire set. The entire process is visualized in Figure \ref{fig:imdp_pipeline}.

\noindent The pipeline consists of three main stages: (1) Pre-filtering and Initial Scoring, (2) Iterative Score Refinement, and (3) Final Filtering via Automated Thresholding.

\subsection{Stage 1: Candidate Pre-filtering and Initial Scoring}

This initial stage aims to drastically reduce non-linguistic noise and establish a baseline score for each plausible candidate.

\subsubsection{Hard Pre-filtering}

First, we apply a series of deterministic filters to the raw input list $C_{raw}$. A token $t \in C_{raw}$ is discarded if it:

\begin{enumerate}
    \item Contains symbols from a non-target script (e.g., Cyrillic in a Latin-based list). We define a valid character set $\Sigma$ for each language (e.g., [a-záéíóöőúüű] for Hungarian).
    \item Contains any non-alphabetic characters (e.g., numbers, punctuation, URLs), excluding initial/final hyphens used to mark affixes.
    \item Is a proper noun or acronym (heuristic: starts with a capital letter or consists of multiple uppercase letters).
    \item Is excessively long ($|t| > 30$) or too short ($|t| < min\_length$), unless $t$ is a single character present in a language-specific whitelist of valid one-character morphemes W.
\end{enumerate}

\subsubsection{Type-support Filtering}

To filter out typographical errors and other singleton noise, we apply a "type-support" criterion to the remaining set of candidates $C'$. A candidate $t \in C'$ is kept only if it appears as a substring in at least $m$ other unique candidates in $C'$. This ensures that we only consider patterns that are structurally recurrent within the dataset itself.
$support(t) = |\{c \in C' | t$ is a substring of $c\}|$
We retain $t$ if $support(t) \geq m$ (we use $m=3$). The resulting set is our final candidate pool $C$.

\subsubsection{Initial Atomicity Scoring}

Each surviving candidate $t \in C$ is assigned an initial Atomicity Score $S_0(t)$. This score is based on the MDL-inspired principle that, all else being equal, shorter forms are more likely to be fundamental morphemic units. The score is defined as the inverse of the token's length:
$S_0(t) = \frac{1}{|t|}$, where $|t|$ is the number of characters in $t$.

\subsection{Stage 2: Iterative Score Refinement}

This is the core of our method. We iteratively refine the Atomicity Scores until they converge. In each iteration $k+1$, the score of every token $t \in C$ is re-calculated based on its "explainability" by other tokens in the set.

\subsubsection{Optimal Decomposition and Best Explanation Power (BEP)}

For each token $t$, we find its optimal decomposition into a sequence of smaller tokens $(m_1, m_2, ..., m_n)$ where each $m_i \in C$. The optimal decomposition is the one that maximizes the sum of the scores of its constituents (taken from the previous iteration, $S_k$). We find this maximum sum using a dynamic programming algorithm and term it the Best Explanation Power, $BEP_k(t)$.
\[
\mathrm{BEP}_{k}(t)=
\max_{\substack{t = m_1\cdots m_n \\ n\ge 2}}
\sum_{i=1}^{n} S_{k}(m_i).
\]
The search space for decompositions is constrained by two rules:

\begin{enumerate}
    \item \textbf{Multi-component:} The algorithm considers segmentations into any number of parts, not just two.
    \item \textbf{Degeneracy Prevention:} Segments of length 1 are only considered if they are in the whitelist $W$.
\end{enumerate}

\subsubsection{Score Update Rule}

The new score $S_{k+1}(t)$ is calculated by comparing the token's own score with its explainability. A token is penalized only if the "evidence" for it being composite ($BEP_k(t)$) is stronger than the evidence for it being an atom ($S_k(t)$).

\[
S_{k+1}(t) =
\begin{cases}
  S_k(t), & \text{if } \mathrm{BEP}_k(t) \le S_k(t),\\[4pt]
  \dfrac{S_0(t)}{1 + \mathrm{BEP}_k(t)}, & \text{if } \mathrm{BEP}_k(t) > S_k(t).
\end{cases}
\]

\noindent This update rule creates a competitive dynamic where atomic morphemes retain high scores, while composite words are iteratively penalized towards zero.

\subsubsection{Convergence}

The iterative process continues until the system reaches a stable state. We define convergence as the point where the maximum absolute change in any token's score between two consecutive iterations falls below a small threshold 
\[
\max_{t \in \mathcal{C}} \bigl| S_{k+1}(t) - S_k(t) \bigr| < \varepsilon
\]
We use $\varepsilon = 1e-7$ and a safeguard limit of $max\_iterations = 100$.

\subsection{Stage 3: Final Filtering via Automated Thresholding}

After the scores converge, the final distribution of scores typically shows a heavy concentration of composite candidates at very low scores, while atomic candidates retain higher scores. To automatically and reproducibly determine a separation threshold between these groups, we employ Otsu's method \cite{otsu1979threshold}. Originally developed for image processing to separate foreground from background, this algorithm finds an optimal threshold $\tau$ for a distribution by maximizing the inter-class variance between the two resulting classes (in our case, "atomic" vs. "composite"). This data-driven approach avoids manual parameter tuning and adapts to the specific score distribution of each dataset.

All tokens $t$ with a final score $S_{final}(t) >= \tau$ are classified as atomic and form our final, high-purity morpheme lexicon.

\begin{table}[h]
\centering
\begin{tabular}{lrrcc}
\hline
\textbf{Lang} & \textbf{Initial} & \textbf{Atomic} & \textbf{Reduct} & \textbf{Reduct} \\
\textbf{} & \textbf{Cands} & \textbf{Morphs} & \textbf{\%} & \textbf{Factor} \\
\hline
Fin   & 499,647 & 3,850 & 99.23\% & 129.8x \\
Est  & 281,256 & 5,705 & 97.97\% & 49.3x \\
Hung & 103,317 & 3,189 & 96.91\% & 32.4x \\
\hline
\end{tabular}
\caption{Efficiency of the IMDP pipeline in cleaning and reducing morpheme candidate lists.}
\label{tab:pipeline_efficiency}
\end{table}

\section{Experimental Setup}

To evaluate the impact of vocabulary size on morphological coverage, we conducted a systematic analysis for three Uralic languages: Finnish, Hungarian, and Estonian. Our experimental setup consists of three main stages: creating the reference morphemes, training the tokenizers, and defining the evaluation metrics.

\subsection{Data}

Our methodology requires two types of data for each language: a raw list of morpheme candidates for cleaning and a large text corpus for tokenizer training.

\begin{enumerate}
    \item \textbf{Morpheme Candidate Lists:} The initial "dirty" lists of candidates were constructed from authoritative, open-source spell-checking dictionaries based on the Hunspell framework\footnote{\url{https://hunspell.github.io/}}. For Hungarian and Estonian, we utilized the comprehensive dictionaries curated by The LibreOffice Project\footnote{\url{https://github.com/LibreOffice/dictionaries}}. For Finnish, which requires special handling of compounds, we used the dedicated dictionary from the hunspell-fi project\footnote{\url{https://github.com/fginter/hunspell-fi}}. For each language, the full set of unique stems (from.dicfiles) and affixes (from.afffiles) was merged to create a comprehensive but structurally noisy candidate list, which serves as the input to our IMDP pipeline. This approach of leveraging widely available dictionary resources provides a practical starting point for morphological analysis.
    \item \textbf{Text Corpora:} For training the BPE tokenizers, we used large, pre-processed corpora derived from Wikipedia snapshots\footnote{\url{https://dumps.wikimedia.org}}. Our choice of data source and preprocessing methodology aligns with previous work on creating specialized Uralic tokenizers \cite{chelombitko-komissarov-2024-specialized}, ensuring a comparable basis for our analysis. It is critical to emphasize that these corpora were used exclusively for training the BPE tokenizers and were not used in any stage of our morpheme list refinement pipeline, thus preserving the corpus-free nature of the IMDP method.

\end{enumerate}

\subsection{Reference Lexicon Creation}

For each of the three languages, we applied our Iterative Morphological Decomposition Pipeline (IMDP), as described in Section 3, to the corresponding raw candidate list. The pipeline was configured with the following parameters: a minimum morpheme length $min\_length=1$, a minimum type-support $m=3$, and a convergence threshold $\varepsilon = 1e-7$. The process was run until convergence. The final filtering was performed using the automatically determined Otsu threshold \cite{otsu1979threshold}. This procedure yielded three high-purity reference morpheme lexicons ($G_{fin}$, $G_{hun}$, $G_{est}$), the statistics of which are summarized in Table \ref{tab:pipeline_efficiency}.

\subsection{Tokenizer Training}

Using the tokenizers library\footnote{\url{https://github.com/huggingface/tokenizers}} and SentencePiece \cite{kudo2018sentencepiece} for comparison, we trained a series of Byte-Pair Encoding (BPE) tokenizers for each language from scratch. The tokenizers were trained on the respective Wikipedia corpora. To analyze the effect of vocabulary size, we trained separate models for a range of vocabulary sizes k, starting from 8,000 and up to 256,000 ($k \in $ \{8k, 16k, 32k, 40k, 50k, 64k, 80k, 100k, 128k, 150k, 180k, 200k, 220k, 240k, 256k\}). All tokenizers were trained with a $min\_frequency$ of $2$ for merges.

\subsection{Evaluation Metrics}

To provide a nuanced and rigorous evaluation of tokenizer quality, we must account for the fundamental trade-off between morphological coverage and over-segmentation. A tokenizer that perfectly represents all morphemes (high coverage) but also excessively splits common words is not optimal. To capture this balance in a single, unified score, we introduce the Integrated Performance Score (IPS).

The IPS models this trade-off geometrically. We consider a 2D space where the ideal tokenizer resides at the point (Coverage=1, OverSplit=0). The IPS of any real tokenizer is its normalized Euclidean distance from this ideal point, scaled to a [0, 1] range where 1 is perfect.

First, we define the two core components:

\begin{enumerate}
    \item \textbf{Lexical Morpheme Coverage (LMC):} The fraction of atomic morphemes from our reference lexicon $G$ that are perfectly represented as a single token in the tokenizer's vocabulary $V_k$. This measures the tokenizer's lexical "knowledge" of fundamental morphological units. 
    \[
    \mathrm{LMC} \;=\; \frac{\bigl|\{\, m \in G \;\mid\; m \in V_k \,\}\bigr|}{|G|}.
    \]
    \item \textbf{Over-split Rate (OSR):} The fraction of morphemes from G that the tokenizer fails to represent as single tokens, thus always splitting them into multiple pieces.
    \[
    \mathrm{OSR}=
    \frac{
    \left|\left\{\,m\in M \,\middle|\,
    \substack{
    m \text{ occurs in $\geq$1 word}\\
    m \text{ never as a single token}
    }\right\}\right|
    }{
    \left|\left\{\,m\in M \,\middle|\, m \text{ in $\geq$1 word} \right\}\right|
    }.
    \]

\end{enumerate}

\noindent From these, the Integrated Performance Score (IPS) is calculated as: \\

$IPS = 1 - (\frac{\sqrt{(1 - LMC)^2 + OSR^2}}{\sqrt{2}})$ \\

\noindent This single metric allows for a clear and direct comparison of tokenizers across different vocabulary sizes. A higher IPS indicates a better balance between representing morphemes and avoiding excessive fragmentation. Our final analysis of optimal vocabulary sizes is based on identifying the "elbow point" on the IPS vs. vocabulary size curve.

\section{Results and Analysis}

\begin{figure}[h]
    \centering
    \includegraphics[width=1.0\linewidth]{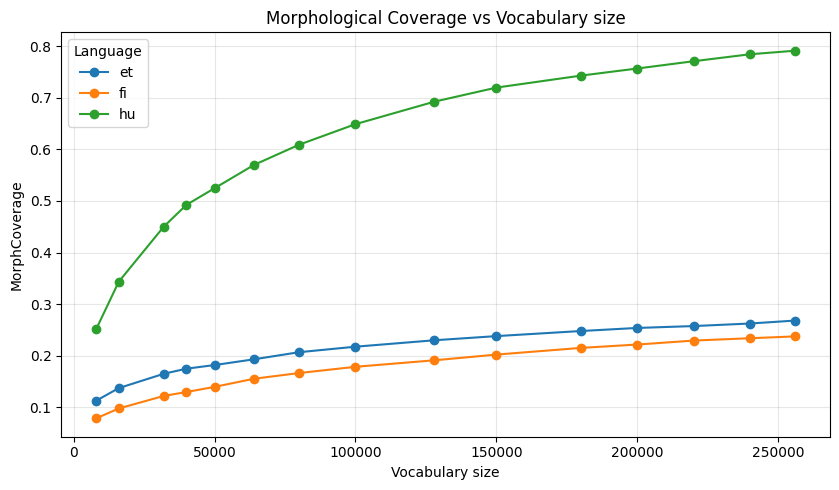}
    \caption{Lexical Morpheme Coverage (LMC) across different vocabulary sizes (k). LMC represents the percentage of reference morphemes found as single, complete tokens in the tokenizer's vocabulary.}
    \label{fig:morph_coverage_all}
\end{figure}

\begin{figure}[h]
    \centering
    \includegraphics[width=1.0\linewidth]{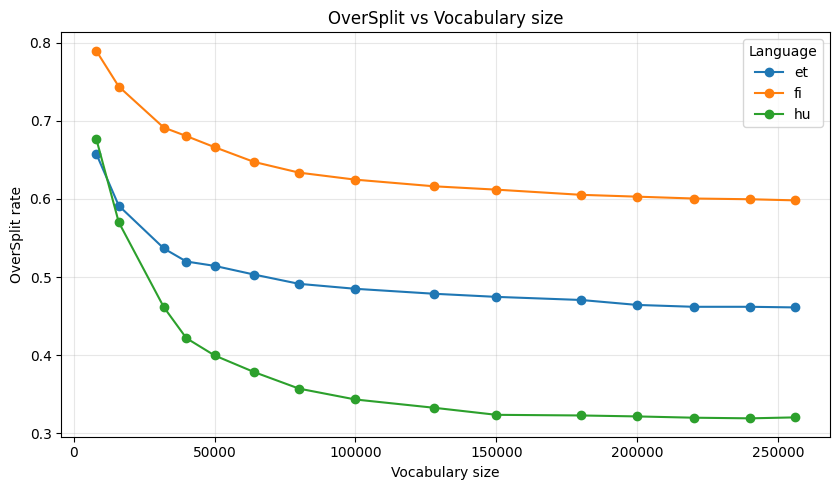}
    \caption{Over-Split Rate (OSR) as a function of vocabulary size (k). OSR denotes the fraction of reference morphemes that occur in words but never appear as a single token in any tokenization.}
    \label{fig:oversplit_all}
\end{figure}

Our experiment yielded clear and significant patterns regarding the relationship between tokenizer vocabulary size and morphological performance. To capture the fundamental trade-off between coverage and over-segmentation, we analyzed the Integrated Performance Score (IPS) for each language. The resulting IPS curves for Estonian (Figure \ref{fig:ips_et}), Finnish (Figure \ref{fig:ips_fi}), and Hungarian (Figure \ref{fig:ips_hu}) clearly show the performance profile for each language. Supplementary details on the component metrics (LMC and OSR) available in Figures \ref{fig:morph_coverage_all} and \ref{fig:oversplit_all}.

\subsection{General Observation: A Clear Trade-off Profile}

The IPS curves for all three languages exhibit a classic logarithmic growth pattern, demonstrating the law of diminishing returns. The score increases rapidly for smaller vocabulary sizes, indicating that initial additions to the vocabulary are highly efficient at capturing morphological structure. However, the rate of improvement progressively slows, showing that ever-larger vocabularies provide only marginal gains at a significant cost to model size. This confirms that a "sweet spot" or an optimal range exists for each language.

\subsection{Cross-Linguistic Analysis: Three Distinct Performance Tiers}

The results reveal three distinct performance tiers, highlighting the varying degrees to which standard BPE can model the morphology of these languages.

\begin{figure}[h]
    \centering
    \includegraphics[width=1.0\linewidth]{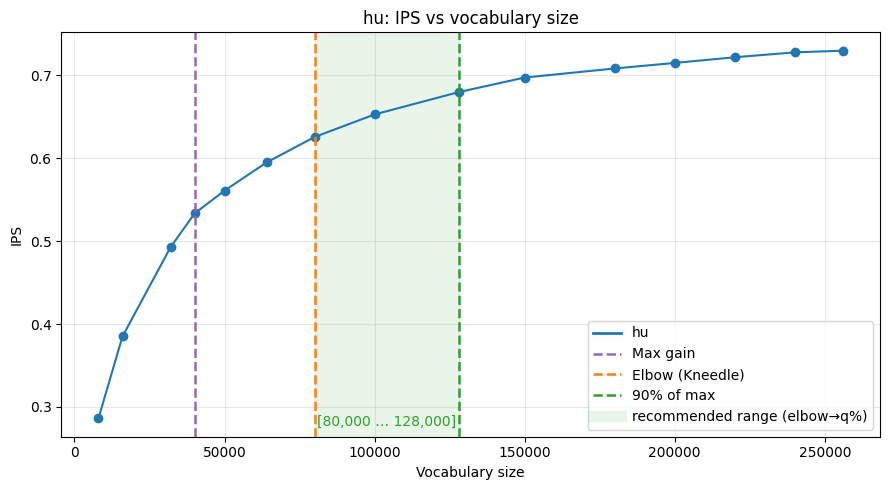}
    \caption{IPS vs. vocabulary size (k) for \textbf{Hungarian}. Hungarian shows the most consistent improvement in IPS, reflecting its comparatively transparent agglutinative structure with fewer morphophonological alternations. The elbow point is at 80k, and the 90\% quality threshold at 128k, yielding a recommended range of 80k–128k.}
    \label{fig:ips_hu}
\end{figure}

\begin{enumerate}
    \item \textbf{Hungarian (hu):} As shown in Figure \ref{fig:ips_hu}, Hungarian demonstrates by far the best performance. Its IPS curve starts at 0.29 and rises sharply, reaching a maximum of 0.73. This high score suggests that BPE is reasonably effective at learning the statistical regularities of Hungarian morphology.
    \item \textbf{Estonian (et):} Estonian occupies the middle tier, with its IPS curve depicted in Figure \ref{fig:ips_et}. The score starts at 0.22 and reaches a maximum of 0.39. While better than Finnish, this score indicates that less than 40\% of the "ideal" tokenizer performance is achieved, even with a large vocabulary.
    \item \textbf{Finnish (fi):} Figure \ref{fig:ips_fi} illustrates the most challenging profile for Finnish. With a maximum IPS of only 0.31, the results quantitatively demonstrate that standard BPE is fundamentally ill-suited for capturing the complexities of Finnish morphology.
\end{enumerate}

\subsection{Identifying the Optimal Vocabulary Range (k*)}

\begin{figure}[h]
    \centering
    \includegraphics[width=1.0\linewidth]{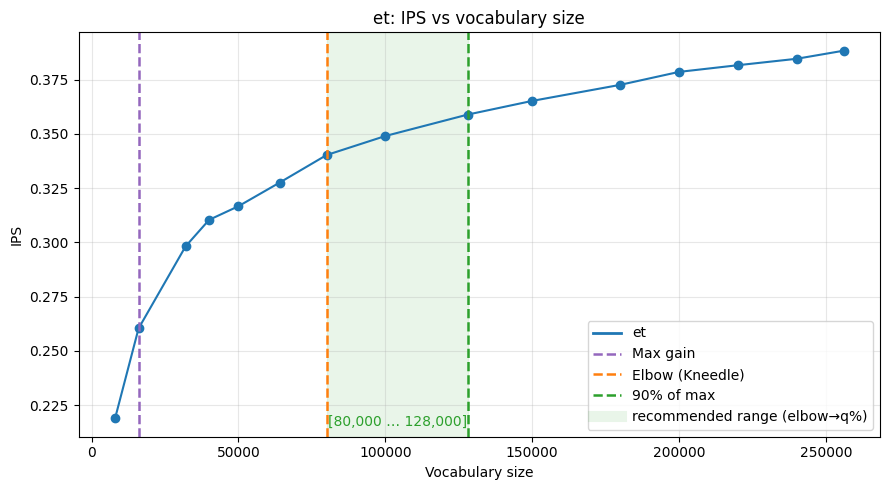}
    \caption{IPS vs. vocabulary size (k) for \textbf{Estonian}. While the overall pattern of diminishing returns is similar to Hungarian, the lower IPS plateau indicates reduced learnability due to Estonian’s extensive morphophonological alternations, which obscure orthographic morpheme boundaries. The recommended range remains 80k–128k.}
    \label{fig:ips_et}
\end{figure}

\begin{figure}[h]
    \centering
    \includegraphics[width=1.0\linewidth]{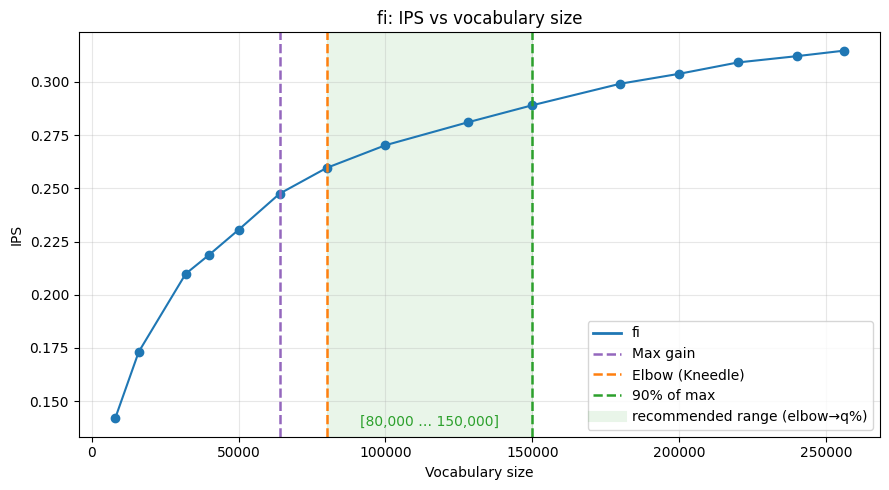}
    \caption{IPS vs. vocabulary size (k) for \textbf{Finnish}. Finnish exhibits the lowest IPS plateau, consistent with its rich system of consonant gradation and stem alternations, which make orthographic segmentation less stable for BPE. The elbow is at 80k, while 90\% of the maximum IPS is reached at 150k, suggesting a recommended range of 80k–150k.}
    \label{fig:ips_fi}
\end{figure}

To determine a practical and effective vocabulary size, we define a recommended range for k*. The lower bound of this range is the "elbow" point (k\_elbow), identified by the Kneedle algorithm \cite{satopaa2011finding}, which marks the point of diminishing returns. The upper bound is the 90\% quality point (k\_q90), where 90\% of the maximum observed IPS is achieved.
As shown in Figures \ref{fig:ips_hu}, \ref{fig:ips_fi}, \ref{fig:ips_et}, and summarized in Table \ref{tab:summarized_results}, this analysis leads to the following recommendations:

\begin{table*}[h]
\centering
\begin{tabular}{lcccc}
\hline
\textbf{Lang} & \textbf{\makecell{Max Gain Point \\ (k\_gain)}} & \textbf{\makecell{Elbow Point \\ (k\_elbow)}} & \textbf{\makecell{90\% Quality Point \\ (k\_q90)}} & \textbf{\makecell{Recommend\\ k* Range}} \\
\hline
Hung & 40,000 & 80,000 & 128,000 & 80k -- 128k \\
Est  & 16,000 & 80,000 & 128,000 & 80k -- 128k \\
Fin   & 64,000 & 80,000 & 150,000 & 80k -- 150k \\
\hline
\end{tabular}
\caption{Key points on the IPS curve for determining the optimal vocabulary range.}
\label{tab:summarized_results}
\end{table*}

\begin{enumerate}
    \item \textbf{Hungarian (hu):} The IPS curve for Hungarian (Figure \ref{fig:ips_hu}) shows a clear optimal range between k=80,000 and k=128,000. The elbow is found at 80k, and 90\% of the maximum performance is reached at 128k. As visualized on the plot, expanding the vocabulary beyond this range yields only minimal performance gains.
    \item \textbf{Estonian (et):} For Estonian (Figure \ref{fig:ips_et}), the recommended range is also k=80,000 to k=128,000. Similar to Hungarian, the elbow is at 80k and the 90\% quality mark is at 128k, establishing this as the zone of best compromise between performance and size.
    \item \textbf{Finnish (fi):} The analysis for Finnish (Figure \ref{fig:ips_fi}) indicates a need for a larger vocabulary. The elbow is at k=80,000, but to achieve 90\% of the (albeit low) maximum performance, a vocabulary of k=150,000 is required. This suggests that for Finnish, the optimal range is k=80,000 to k=150,000, reflecting the language's high morphological complexity.
\end{enumerate}

\noindent These findings provide a quantitative foundation for the critical decision of vocabulary sizing, transforming it from a heuristic-based choice into a principled optimization problem. Complete numerical results for all evaluated vocabulary sizes are provided in Appendix~A (Table \ref{tab:full_results}) for reference.

\section{Conclusion}
\label{sec:bibtex}

In this work, we addressed the dual challenge of creating high-purity morphological resources in a corpus-free setting and using them to evaluate subword tokenizers for Uralic languages. We introduced SampoNLP, a toolkit featuring a novel pipeline based on "MDL-inspired Self-Referential Atomicity Scoring," which successfully refines noisy candidate lists into clean morpheme lexicons.

Applying these lexicons, our systematic evaluation of BPE tokenizers yielded two key findings. First, we provide an empirically-grounded recommendations for optimal vocabulary sizes, identifying a range of 80k-128k for Hungarian and Estonian, and 80k-150k for Finnish, as the most effective trade-off between performance and model size. Second, our results quantitatively demonstrate the severe limitations of standard BPE for highly agglutinative languages like Finnish, where performance plateaus at a strikingly low level.

This study confirms that while vocabulary size optimization is a crucial step, it is not a panacea. We release our SampoNLP library and the generated morpheme lists to the community to facilitate reproducible research and encourage the development of more morphologically-aware tokenization methods for the Uralic language family.

\section*{Discussion}

Our results yield two key insights. First, the effectiveness of BPE varies dramatically by language: while Hungarian achieves a high IPS (max $\sim$0.73), the low scores for Finnish ($\sim$0.31) and Estonian ($\sim$0.39) quantitatively demonstrate the algorithm's fundamental limitations for these highly agglutinative languages. Second, for all languages, an empirically identifiable "sweet spot" for vocabulary size exists, beyond which performance gains diminish. Here, “optimality” is understood as morphological sufficiency - the point at which the tokenizer captures the productive structure of a language with minimal redundancy. This notion is intrinsic by design, offering a language-level criterion rather than task-specific optimization.

We acknowledge the limitations of our approach. The IPS metric abstracts away qualitative segmentation differences - a necessary compromise for scalability. Our use of clean, standardized corpora also isolates the variable of vocabulary size but does not reflect the noise of real-world data. These aspects represent clear avenues for future work.

While our method produces a refined set of recurrent sub-lexical units, we do not claim full linguistic morpheme correctness. The IMDP segmentation is orthographic and self-referential in nature, providing a practical approximation rather than a phonologically grounded morphological analysis.

In conclusion, our findings suggest that while optimizing k* is a crucial step, it may be insufficient for languages like Finnish. The low performance ceiling for BPE underscores the need for morphologically-aware tokenization methods. We believe our SampoNLP toolkit and the generated lexicons provide the community with a reproducible benchmark to develop and test such new strategies.



\bibliography{custom}

@inproceedings{chelombitko-komissarov-2024-specialized,
    title = "Specialized Monolingual {BPE} Tokenizers for {U}ralic Languages Representation in Large Language Models",
    author = "Chelombitko, Iaroslav  and
      Komissarov, Aleksey",
    editor = {H{\"a}m{\"a}l{\"a}inen, Mika  and
      Pirinen, Flammie  and
      Macias, Melany  and
      Crespo Avila, Mario},
    booktitle = "Proceedings of the 9th International Workshop on Computational Linguistics for Uralic Languages",
    month = nov,
    year = "2024",
    address = "Helsinki, Finland",
    publisher = "Association for Computational Linguistics",
    url = "https://aclanthology.org/2024.iwclul-1.11/",
    pages = "89--95"
}

@misc{chelombitko2024qtok,
      title   = {Qtok: A Comprehensive Framework for Evaluating Multilingual Tokenizer Quality in Large Language Models}, 
      author  = {Iaroslav Chelombitko and Egor Safronov and Aleksey Komissarov},
      year    = {2024},
      eprint  = {2410.12989},
      archivePrefix = {arXiv},
      primaryClass = {cs.CL},
      url     = {https://arxiv.org/abs/2410.12989}, 
}

@misc{popova2025repeatsdrivevocabularybytepair,
      title   = {When repeats drive the vocabulary: a Byte-Pair Encoding analysis of T2T primate genomes}, 
      author  = {Marina Popova and Iaroslav Chelombitko and Aleksey Komissarov},
      year    = {2025},
      eprint  = {2505.08918},
      archivePrefix = {arXiv},
      primaryClass = {q-bio.GN},
      url     = {https://arxiv.org/abs/2505.08918}, 
}

@inproceedings{sennrich-etal-2016-neural,
    title = "Neural Machine Translation of Rare Words with Subword Units",
    author = "Sennrich, Rico  and
      Haddow, Barry  and
      Birch, Alexandra",
    editor = "Erk, Katrin  and
      Smith, Noah A.",
    booktitle = "Proceedings of the 54th Annual Meeting of the Association for Computational Linguistics (Volume 1: Long Papers)",
    month = aug,
    year = "2016",
    address = "Berlin, Germany",
    publisher = "Association for Computational Linguistics",
    url = "https://aclanthology.org/P16-1162/",
    doi = "10.18653/v1/P16-1162",
    pages = "1715--1725"
}

@inproceedings{bostrom-durrett-2020-byte,
    title = "Byte Pair Encoding is Suboptimal for Language Model Pretraining",
    author = "Bostrom, Kaj  and
      Durrett, Greg",
    editor = "Cohn, Trevor  and
      He, Yulan  and
      Liu, Yang",
    booktitle = "Findings of the Association for Computational Linguistics: EMNLP 2020",
    month = nov,
    year = "2020",
    address = "Online",
    publisher = "Association for Computational Linguistics",
    url = "https://aclanthology.org/2020.findings-emnlp.414/",
    doi = "10.18653/v1/2020.findings-emnlp.414",
    pages = "4617--4624"
}

@article{creutz-lagus-2007-unsupervised,
author = {Creutz, Mathias and Lagus, Krista},
title = {Unsupervised models for morpheme segmentation and morphology learning},
year = {2007},
issue_date = {January 2007},
publisher = {Association for Computing Machinery},
address = {New York, NY, USA},
volume = {4},
number = {1},
issn = {1550-4875},
url = {https://doi.org/10.1145/1187415.1187418},
doi = {10.1145/1187415.1187418},
journal = {ACM Trans. Speech Lang. Process.},
month = feb,
articleno = {3},
numpages = {34}
}

@inproceedings{pirinen-2015-omorfi,
    title = "{O}morfi {---} Free and open source morphological lexical database for {F}innish",
    author = "Pirinen, Tommi A",
    editor = "Megyesi, Be{\'a}ta",
    booktitle = "Proceedings of the 20th Nordic Conference of Computational Linguistics ({NODALIDA} 2015)",
    month = may,
    year = "2015",
    address = "Vilnius, Lithuania",
    publisher = {Link{\"o}ping University Electronic Press, Sweden},
    url = "https://aclanthology.org/W15-1844/",
    pages = "313--315"
}

@article{riessler-2022-uralic,
  title     = {UralicNLP: An NLP Library for Uralic Languages},
  author    = {Hämäläinen, Mika},
  journal   = {Journal of Open Source Software},
  volume    = {4},
  number    = {37},
  year      = {2019},
  doi       = {10.21105/joss.01345},
  url       = {https://doi.org/10.21105/joss.01345}
}

@inproceedings{hamalainen-2021-neural,
    title = "Neural Morphology Dataset and Models for Multiple Languages, from the Large to the Endangered",
    author = {H{\"a}m{\"a}l{\"a}inen, Mika  and
      Partanen, Niko  and
      Rueter, Jack  and
      Alnajjar, Khalid},
    editor = "Dobnik, Simon  and
      {\O}vrelid, Lilja",
    booktitle = "Proceedings of the 23rd Nordic Conference on Computational Linguistics (NoDaLiDa)",
    month = may # " 31--2 " # jun,
    year = "2021",
    address = "Reykjavik, Iceland (Online)",
    publisher = {Link{\"o}ping University Electronic Press, Sweden},
    url = "https://aclanthology.org/2021.nodalida-main.17/",
    pages = "166--177"
}

@article{rissanen-1978-modeling,
  author  = {Jorma Rissanen},
  title   = {Modeling by Shortest Data Description},
  journal = {Automatica},
  volume  = {14},
  number  = {5},
  pages   = {465--471},
  year    = {1978},
}

@inproceedings{kudo-2018-subword,
    title = "Subword Regularization: Improving Neural Network Translation Models with Multiple Subword Candidates",
    author = "Kudo, Taku",
    editor = "Gurevych, Iryna  and
      Miyao, Yusuke",
    booktitle = "Proceedings of the 56th Annual Meeting of the Association for Computational Linguistics (Volume 1: Long Papers)",
    month = jul,
    year = "2018",
    address = "Melbourne, Australia",
    publisher = "Association for Computational Linguistics",
    url = "https://aclanthology.org/P18-1007/",
    doi = "10.18653/v1/P18-1007",
    pages = "66--75"
}

@misc{mielke-etal-2021-results,
      title={Between words and characters: A Brief History of Open-Vocabulary Modeling and Tokenization in NLP}, 
      author={Sabrina J. Mielke and Zaid Alyafeai and Elizabeth Salesky and Colin Raffel and Manan Dey and Matthias Gallé and Arun Raja and Chenglei Si and Wilson Y. Lee and Benoît Sagot and Samson Tan},
      year={2021},
      eprint={2112.10508},
      archivePrefix={arXiv},
      primaryClass={cs.CL},
      url={https://arxiv.org/abs/2112.10508}, 
}

@inproceedings{arkhangelskiy-2020-corpus,
  title     = {Corpora of social media in minority {U}ralic languages},
  author    = {Arkhangelskiy, Timofey},
  booktitle = {Proceedings of the Fifth International Workshop on Computational Linguistics for Uralic Languages},
  month     = jan,
  year      = {2019},
  address   = {Tartu, Estonia},
  publisher = {Association for Computational Linguistics},
  url       = {https://aclanthology.org/W19-0311/},
  doi       = {10.18653/v1/W19-0311},
  pages     = {125--140}
}

@article{trosterud-2012-uralic,
  title     = {Uralic Language Identification (ULI) 2020 shared task: Wanca 2017 web corpora for Uralic languages},
  author    = {Jauhiainen, T. and Linden, Krister and Partanen, Niko and …},
  journal   = {Proceedings of the VarDial Workshop at LREC 2020},
  year      = {2020},
  url       = {https://aclanthology.org/2020.vardial-1.16.pdf}
}

@misc{virtanen-etal-2019-multilingual,
      title={Multilingual is not enough: BERT for Finnish}, 
      author={Antti Virtanen and Jenna Kanerva and Rami Ilo and Jouni Luoma and Juhani Luotolahti and Tapio Salakoski and Filip Ginter and Sampo Pyysalo},
      year={2019},
      eprint={1912.07076},
      archivePrefix={arXiv},
      primaryClass={cs.CL},
      url={https://arxiv.org/abs/1912.07076}, 
}

@inproceedings{acs-2019-tokenization,
    title = "Analyzing Cognitive Plausibility of Subword Tokenization",
    author = "Beinborn, Lisa  and
      Pinter, Yuval",
    editor = "Bouamor, Houda  and
      Pino, Juan  and
      Bali, Kalika",
    booktitle = "Proceedings of the 2023 Conference on Empirical Methods in Natural Language Processing",
    month = dec,
    year = "2023",
    address = "Singapore",
    publisher = "Association for Computational Linguistics",
    url = "https://aclanthology.org/2023.emnlp-main.272/",
    doi = "10.18653/v1/2023.emnlp-main.272",
    pages = "4478--4486"
}

@article{otsu1979threshold,
  author    = {Otsu, Nobuyuki},
  title     = {A Threshold Selection Method from Gray-Level Histograms},
  journal   = {IEEE Transactions on Systems, Man, and Cybernetics},
  year      = {1979},
  volume    = {9},
  number    = {1},
  pages     = {62--66},
  doi       = {10.1109/TSMC.1979.4310076}
}

@inproceedings{satopaa2011finding,
  title={Finding a “Kneedle” in a Haystack: Detecting Knee Points in System Behavior},
  author={Satopää, Ville and Albrecht, Joshua and Irwin, David and Raghavan, Barath},
  booktitle={Proceedings of the 31st International Conference on Distributed Computing Systems Workshops},
  pages={166--171},
  year={2011},
  publisher={IEEE},
  doi={10.1109/ICDCSW.2011.20}
}

@inproceedings{kudo2018sentencepiece,
  title     = {SentencePiece: A simple and language independent subword tokenizer and detokenizer for Neural Text Processing},
  author    = {Kudo, Taku and Richardson, John},
  booktitle = {Proceedings of the 2018 Conference on Empirical Methods in Natural Language Processing: System Demonstrations},
  pages     = {66--71},
  year      = {2018},
  address   = {Brussels, Belgium},
  publisher = {Association for Computational Linguistics},
  url       = {https://aclanthology.org/D18-2012}
}

@article{feng2004accessor,
author = {Chen, Kang and Deng, Xiaotie and Zheng, Weimin},
year = {2004},
month = {03},
pages = {75-93},
title = {Accessor Variety Criteria for Chinese Word Extraction},
volume = {30},
journal = {Computational Linguistics},
doi = {10.1162/089120104773633394}
}

@inproceedings{rust-etal-2021-good,
    title = "How Good is Your Tokenizer? On the Monolingual Performance of Multilingual Language Models",
    author = "Rust, Phillip  and
      Pfeiffer, Jonas  and
      Vuli{\'c}, Ivan  and
      Ruder, Sebastian  and
      Gurevych, Iryna",
    editor = "Zong, Chengqing  and
      Xia, Fei  and
      Li, Wenjie  and
      Navigli, Roberto",
    booktitle = "Proceedings of the 59th Annual Meeting of the Association for Computational Linguistics and the 11th International Joint Conference on Natural Language Processing (Volume 1: Long Papers)",
    month = aug,
    year = "2021",
    address = "Online",
    publisher = "Association for Computational Linguistics",
    url = "https://aclanthology.org/2021.acl-long.243/",
    doi = "10.18653/v1/2021.acl-long.243",
    pages = "3118--3135"
}

@article{gerz-etal-2018-language,
    title = "Language Modeling for Morphologically Rich Languages: Character-Aware Modeling for Word-Level Prediction",
    author = "Gerz, Daniela  and
      Vuli{\'c}, Ivan  and
      Ponti, Edoardo  and
      Naradowsky, Jason  and
      Reichart, Roi  and
      Korhonen, Anna",
    editor = "Lee, Lillian  and
      Johnson, Mark  and
      Toutanova, Kristina  and
      Roark, Brian",
    journal = "Transactions of the Association for Computational Linguistics",
    volume = "6",
    year = "2018",
    address = "Cambridge, MA",
    publisher = "MIT Press",
    url = "https://aclanthology.org/Q18-1032/",
    doi = "10.1162/tacl_a_00032",
    pages = "451--465"
}

\appendix

\section{Appendix}
\label{sec:appendix}



\begin{table*}[h]
\centering
\begin{tabular}{llrrc} 
\hline
\textbf{Language} & \textbf{\makecell{Vocabulary \\ Size (k)}} & \textbf{\makecell{Total \\ Morphemes}} & \textbf{\makecell{Morpheme \\ Coverage \%}} & \textbf{\makecell{Over-Split \\ Rate \%}} \\
\hline
Estonian  & 8,000   & 5,705 & 11.27\% & 65.79\% \\
Estonian  & 16,000  & 5,705 & 13.71\% & 59.13\% \\
Estonian  & 32,000  & 5,705 & 16.49\% & 53.65\% \\
Estonian  & 40,000  & 5,705 & 17.48\% & 51.98\% \\
Estonian  & 50,000  & 5,705 & 18.18\% & 51.43\% \\
Estonian  & 64,000  & 5,705 & 19.30\% & 50.32\% \\
Estonian  & 80,000  & 5,705 & 20.68\% & 49.13\% \\
Estonian  & 100,000 & 5,705 & 21.74\% & 48.49\% \\
Estonian  & 128,000 & 5,705 & 22.99\% & 47.86\% \\
Estonian  & 150,000 & 5,705 & 23.79\% & 47.46\% \\
Estonian  & 180,000 & 5,705 & 24.78\% & 47.06\% \\
Estonian  & 200,000 & 5,705 & 25.38\% & 46.43\% \\
Estonian  & 220,000 & 5,705 & 25.74\% & 46.19\% \\
Estonian  & 240,000 & 5,705 & 26.23\% & 46.19\% \\
Estonian  & 256,000 & 5,705 & 26.81\% & 46.11\% \\
\hline
Finnish   & 8,000   & 3,850 & 7.85\%  & 78.96\% \\
Finnish   & 16,000  & 3,850 & 9.76\%  & 74.37\% \\
Finnish   & 32,000  & 3,850 & 12.20\% & 69.13\% \\
Finnish   & 40,000  & 3,850 & 12.95\% & 68.04\% \\
Finnish   & 50,000  & 3,850 & 13.95\% & 66.62\% \\
Finnish   & 64,000  & 3,850 & 15.53\% & 64.73\% \\
Finnish   & 80,000  & 3,850 & 16.63\% & 63.36\% \\
Finnish   & 100,000 & 3,850 & 17.84\% & 62.46\% \\
Finnish   & 128,000 & 3,850 & 19.11\% & 61.61\% \\
Finnish   & 150,000 & 3,850 & 20.21\% & 61.18\% \\
Finnish   & 180,000 & 3,850 & 21.51\% & 60.52\% \\
Finnish   & 200,000 & 3,850 & 22.16\% & 60.28\% \\
Finnish   & 220,000 & 3,850 & 22.93\% & 60.05\% \\
Finnish   & 240,000 & 3,850 & 23.38\% & 59.95\% \\
Finnish   & 256,000 & 3,850 & 23.73\% & 59.81\% \\
\hline
Hungarian & 8,000   & 3,189 & 25.15\% & 67.72\% \\
Hungarian & 16,000  & 3,189 & 34.34\% & 57.01\% \\
Hungarian & 32,000  & 3,189 & 45.03\% & 46.14\% \\
Hungarian & 40,000  & 3,189 & 49.23\% & 42.17\% \\
Hungarian & 50,000  & 3,189 & 52.46\% & 39.97\% \\
Hungarian & 64,000  & 3,189 & 56.98\% & 37.84\% \\
Hungarian & 80,000  & 3,189 & 60.90\% & 35.72\% \\
Hungarian & 100,000 & 3,189 & 64.88\% & 34.33\% \\
Hungarian & 128,000 & 3,189 & 69.24\% & 33.27\% \\
Hungarian & 150,000 & 3,189 & 71.97\% & 32.37\% \\
Hungarian & 180,000 & 3,189 & 74.29\% & 32.28\% \\
Hungarian & 200,000 & 3,189 & 75.67\% & 32.16\% \\
Hungarian & 220,000 & 3,189 & 77.08\% & 32.00\% \\
Hungarian & 240,000 & 3,189 & 78.43\% & 31.92\% \\
Hungarian & 256,000 & 3,189 & 79.12\% & 32.04\% \\
\bottomrule
\end{tabular}
\caption{Detailed experimental results for BPE tokenizers of varying vocabulary sizes across three Uralic languages. Morpheme Coverage represents the percentage of reference morphemes found in the vocabulary (LMC). Over-Split Rate is the percentage of reference morphemes with support in W that never appear as a single token in any tokenization.}
\label{tab:full_results}
\end{table*}

\end{document}